\documentclass[10pt,letterpaper]{article}
\usepackage[top=0.85in,left=2.75in,footskip=0.75in,marginparwidth=2in]{geometry}

\usepackage[utf8]{inputenc}

\usepackage{cite}

\usepackage{algorithm2e}

\usepackage{nameref,hyperref}

\usepackage[right]{lineno}

\usepackage{microtype}
\DisableLigatures[f]{encoding = *, family = * }

\raggedright
\usepackage{changepage}

\usepackage[aboveskip=1pt,labelfont=bf,labelsep=period,singlelinecheck=off]{caption}

\makeatletter
\renewcommand{\@biblabel}[1]{\quad#1.}
\makeatother

\usepackage{lastpage,fancyhdr,graphicx}
\usepackage{epstopdf}
\fancyheadoffset[L]{2.25in}
\fancyfootoffset[L]{2.25in}

\usepackage{color}

\definecolor{Gray}{gray}{.25}

\usepackage{graphicx}
\usepackage{multirow}

\usepackage{sidecap}

\usepackage{wrapfig}
\usepackage[pscoord]{eso-pic}
\usepackage[fulladjust]{marginnote}
\reversemarginpar

\begin{document}
\vspace*{0.35in}


\begin{flushleft}
{\Large
\textbf\newline{A New Data Representation Based on Training Data \\Characteristics to Extract Drug Named-Entity in\hspace{1.0cm} Medical Text}
}
\newline
\\
Sadikin Mujiono\textsuperscript{1,2},
Mohamad Ivan Fanany\textsuperscript{1},
Chan Basaruddin\textsuperscript{1},
\\
\bigskip
\bf{1} Machine Learning and Computer Vision Laboratory, \\Faculty of Computer Science, Universitas Indonesia\\
\bf{2} Faculty of Computer Science, Universitas Mercu Buana \\
\bigskip
* mujiono.sadikin@mercubuana.ac.id

\end{flushleft}

\providecommand{\keywords}[1]{\textbf{\textit{Keywords---}} #1}

\section*{Abstract}
One essential task in information extraction from the medical corpus is drug name recognition. Compared with text sources come from other domains, the medical text is special and has unique characteristics. In addition, the medical text mining poses more challenges, e.g., more unstructured text, the fast growing of new terms addition, a wide range of name variation for the same drug. The mining is even more challenging due to the lack of labeled dataset sources and external knowledge, as well as multiple token representations for a single drug name that is more common in the real application setting. Although many approaches have been proposed to overwhelm the task, some problems remained with poor F-score performance (less than 0.75). This paper presents a new treatment in data representation techniques to overcome some of those challenges. We propose three data representation techniques based on the characteristics of word distribution and word similarities as a result of word embedding training. The first technique is evaluated with the standard NN model, i.e., MLP (Multi-Layer Perceptrons). The second technique involves two deep network classifiers, i.e., DBN (Deep Belief Networks), and SAE (Stacked Denoising Encoders). The third technique represents the sentence as a sequence that is evaluated with a recurrent NN model, i.e., LSTM (Long Short Term Memory). In extracting the drug name entities, the third technique gives the best F-score performance compared to the state of the art, with its average F-score being 0.8645.
\bigskip

\noindent\keywords{drug name entity, word embedding, MLP, DBN, SAE, LSTM}


\section*{Introduction}
\label{Introduction}
The rapid growth of information technology provides rich text data resources in all areas, including the medical field. An abundant amount of medical text data can be used to obtain valuable information for the benefit of many purposes. The understanding of drug interactions, for example, is an important aspect of manufacturing new medicines or controlling drug distribution in the market. The process to produce a medicinal product is an expensive and complex task. In many recent cases, however, many drugs are withdrawn from the market when it was discovered that the interaction between the drugs is hazardous to health\cite{Sadikin2013}.

Information, or objects extraction, from an unstructured text document, is one of the most challenging studies in the text mining area. The difficulties of text information extraction keeps increasing due to the increasing size of corpora, continuous growth of human's natural language, and the unstructured formatted data \cite{Tang}. Among such valuable information are medical entities such as drug name, compound, and brand; disease name, and their relations, such as drug - drug interaction and drug - compound relation. We need a suitable method to extract such information. To embed those abundant data resources; however, many problems have to be tackled. For example, large data size, unstructured format, choosing the right NLP, and the limitation of annotated datasets.

A more specific and valuable information contained in medical text data is a drug entity (drug name). Drug name recognition is a primary task of medical text data extraction since the drug finding is the essential element in solving other information extraction problems \cite{Zhang2013, Korkontzelos2015}. Among derivative work of drug name extractions are drug-drug interaction \cite{Herrero-zazo2013},  drug adverse reaction \cite{Sampathkumar2014}, or other applications (information retrieval, decision support system, drug development or drug discovery)\cite{Segura-bedmar2008}.

Compared to other NER (Named-Entity Recognition) tasks, such as PERSON, LOCATION, EVENT, or TIME, drug name entity recognition faces more challenges. First, the drug name entities are usually unstructured texts \cite{Keretna2015} where the number of new entities is quickly growing over time. Thus, it is hard to create a dictionary which always includes the entire lexicon and is up to date  \cite{Pal2015}. Second, the naming of the drug also widely varies. The abbreviation and acronym increase the difficulties in determining the concepts referred by the terms. Third, many drug names contain a combination of non-word and word symbols \cite{Liu2015}.Fourth, the other problem in drug name extraction is that a single drug name might be represented by multiple tokens \cite{Grego2013}. Due to the complexity in extracting multiple tokens for drugs, some researchers such as \cite{Bjorne2013} even ignores that case in the MedLine $\&$ DrugBank training with the reason that the multiple tokens drug is only 18\% of all drug names. It is different with another domain, i.e., entity names in the biomedical field are usually longer. Fifth, in some cases, the drug name is a combination of medical and general terms. Sixth, the lack of the labeled dataset is another problem; it has yet to be solved by extracting the drug name entities. 

This paper presents three data representation techniques to extract drug name entities contained in the sentences of medical texts. For the first and the second techniques, we created an instance of the dataset as a tuple, which is formed from 5 vectors of words. In the first technique, the tuple was constructed from all sentences treated as a sequence, whereas in the second technique the tuple is made from each sentence treated as a sequence. The first and second techniques were evaluated with the standard MLP-NN model which is perfomed in the first experiment. In the second experiment, we use the second data representation technique which is also applied to the other NN model i.e. DBN and SAE. The third data representation, which assumes the text as sequential entities, was assessed with the recurrent NN model, LSTM. Those three data representation techniques are based on the word2vec value characteristics, i.e., their cosine and the Euclidean distance between the vectors of words.

In the first and second techniques, we apply three different scenarios to select the most possible words which represent the drug name. The scenarios are based on the characteristics of training data, i.e., drug words distribution that is usually assumed has a smaller frequency of appearance in the dataset sentences. The drug name candidate selections are as follows. In the first case, all test dataset is taken. In the second case, 2/3 of all test dataset is selected. In the third case, $x/y (x < y)$ of the test dataset (where $x$ and $y$ are arbitrary integer numbers) are selected after clustering the test dataset into $y$ clusters. 

In the third experiment, based on the characteristics of the resulting word vectors of the trained word embedding, we formulate a sequence data representation which applied to RNN-LSTM. We used the Euclidian distance of the current input to the previous input as an additional feature besides its vector of words. In this study, the vector of words is provided by word embedding methods proposed by Mikolov \cite{Mikolov2013a}. 

Our main important contributions in this study are:
\begin{enumerate}
    \item The new data representation techniques which do not require any external knowledge nor hand-crafted features.   
    \item The drug extraction techniques based on the words distribution contained in the training data.

\end{enumerate}

Our proposed method is evaluated on DrugBank and MedLine medical open dataset obtained from SemEval 2013 Competition task 9.1, see http://www.cs.york.ac.uk/semeval-2013/task9.html, which is also used by  \cite{Bjorne2013,Grego2013,Ben2015}. The format of both medical texts is in English where some sentences contain drug name entities. In extracting drug entity names from the dataset, our data representation techniques give the best performance with F-score values 0.687 for MLP, 0.6700 for DBN, and 0.682 for SAE, whereas the third technique with LSTM gives the best F-score, i.e., 0.9430. The average F-score of the third technique is 0.8645, i.e., the best performance compared to the other previous methods.        

By applying the data representation techniques, our proposed approach provides at least three advantages:
\begin{enumerate}
    \item The capability to identify multiple tokens as a single name entity. 
    \item The ability to deal with the absence of any external knowledge in certain languages.
    \item No need to construct any additional features, such as characters type identification, orthography feature (lowercase or uppercase identification), or token position.
\end{enumerate}

The rest sections of this paper are organized as follow: Section \ref{related_works} explain some previous works dealing with name entity (and drug name as well) extraction from medical text sources. The framework, approach, and methodology to overcome the challenges of drug name extraction is presented in the Section \ref{method_material}. The section also describes dataset materials and experiment scenarios. Section \ref{result_and_discussion} discusses the experiment results and its analysis while section \ref{Conclusion_Future_Works} explains the achievement, the shortcoming, and the prospects of this study. The section also describes several potential explorations for future research.

\section{Related Works}
\label{related_works}
The entity recognition in a biomedical text is an active research, and many methods have been proposed. For example, Gurinder et al. \cite{Pal2015} summarizes their survey on various entity recognition approaches. The approaches can be categorized into three models: dictionary based, rule based, and learning based methods \cite{Tang,Keretna2015}. A dictionary based approach uses a list of terms (term collection) to assist in predicting which targeted entity will be included in the predicted group. Although their overall precision is more accurate, their recall is poor since they anticipate less new terms. The rule-based approach defines a certain rule which describes such pattern formation surrounding the targeted entity. This rule can be a syntactic term or lexical term. Finally, the learning approach is usually based on statistical data characteristics to build a model using machine learning techniques. The model is capable of automatically learn based on positive, neutral, and negative training data. 

Drug name extraction and their classification are one of the challenges in the Semantic Evaluation Task (SemEval 2013). The best-reported performance for this challenge was 71.5\% in F-score \cite{Segura-Bedmar2013}. Until now the studies to extract drug names still continue and many approaches have been proposed. CRF-based learning is the most common method utilized in the clinical text information extraction. CRF is used by one of the best \cite{Grego2013} participants in SemEval challenges in the clinical text (https://www.cs.york.ac.uk/semeval-2013). As for the use of external knowledge aimed to increase the performance, the author \cite{Grego2013} uses ChEBI (Chemical Entities of Biological Interest), i.e., a dictionary of small molecular entities. The best-achieved performance is 0.57 in F-score (for the overall dataset). 

A hybrid approach model, which combines statistical learning and dictionary based, is proposed by \cite{Segura-bedmar2015}. In their study, the author utilizes word2vec representation, CRF learning model and DINTO, a drug ontology. With this word2vec representation, targeted drug is treated as a current token in a context windows which consists of three tokens on the left and three tokens in the right. Additional features are included in the data representation such as pos tags, lemma in the windows context, an orthography feature as uppercase, lowercase, and mixed cap. The author also used Wikipedia text as an additional resource to perform word2vec representation training. The best F-score value in extracting the drug name provided by the method is 0.72. 

The result of a CRF based active learning, which is applied to NER BIO (Beginning, Inside, Output) annotation token for extracting named entity in the clinical text, is presented in \cite{Chen2015}. The framework of this active learning approach is a sequential process: initial model generation, querying, training, and iteration. The CRF Algorithm BIO approach was also studied by A. Ben et al. \cite{Ben2015}. The features for the CRF algorithm is formulated based on token and linguistics feature and semantic feature. The best F-score achieved by this proposed method is 0.72. 

Korkontzelos et al. studied a combination of aggregated classifier, maximum entropy-multinomial classifier, and handcrafted feature to extract drug entity. \cite{Korkontzelos2015}.They classified drug and non-drug based on the token features formulation such as tokens windows, the current token, and 8 other hand-crafted features.

Another approach for discovering valuable information from clinical text data that adopts event-location extraction model was examined by J, Bjornoe et al. \cite{Bjorne2013}. They use an SVM classifier to predict drug or non-drug entity which is applied to DrugBank dataset. The best performance achieved by their method is a 0.6 in F-score. The drawback of their approach is that it only deals with a single token drug name. 

To overcome the ambiguity problem in NER mined from a medical corpus, a segment representation method has also been proposed by S. Keretna et al. \cite{Keretna2015}. Their approach treats each word as belonging to three classes, i.e., NE, not NE and an ambiguous class. The ambiguity of the class member is determined by identifying whether the word appears in more than one context or not. If so, this word falls into the ambiguous class. After three class segments are found, it is then applied to the classifier learning. Related to their approach, in our previous work, we propose a pattern learning that utilizes the regular expression surrounding drug names and their compounds \cite{Sadikin2014}. The performance of our method is quite good with the average F-score being 0.81 but has a limitation in dealing with more unstructured text data. 

In summarizing the related previous works on drug name entity extraction, we noted some drawbacks which need to be addressed. In general, almost all state of the art methods works based on ad-hoc external knowledge which is not always available. The requirement of the handcrafted feature is another difficult constraint since not all datasets contain such feature. An additional challenge that remained unsolved by the previous works is the problem of multiple tokens representation for a single drug name. This study proposes a new data representation technique to handle those challenges. 

Our proposed method is based only on the data distribution pattern and vector of words characteristics, so there is no need for external knowledge nor additional handcrafted features. To overcome the multiple tokens problem, we propose a new technique which treats a target entity as a set of tokens (a tuple) at once rather than treating the target entity as a single token surrounded by other tokens such as used by \cite{Segura-bedmar2015} or \cite{Bui2014}. By addressing a set of the tokens as a single sample, our proposed method can predict whether a set of tokens is a drug name or not. In our first experiment, we evaluate the first and second data representation techniques and apply MLP learning model. In our second scenario, we choose the the second technique which gave the best result with MLP, and apply it to two different machine learning methods: DBN, and SAE. In our third experiment, we examined the third data representation technique which utilizes the Euclidian distance between successive words in a certain sentence of medical text. The third data representation is then fed into an LSTM model. Based on the resulted F-score value, the second experiment gives the best performance. 

\section{Method $\&$ Material}
\label{method_material}
\subsection{Framework}
\label{framework}

In this study, using the word2vec value characteristics, we conducted three experiments based on different data representation techniques. The first and second experiment examine conventional tuple data representation, whereas the third experiment examines sequence data representation. We describe the organization of these three experiments in this Section. In general, the proposed method to extract drug name entities in this study consists of two main phases. The first phase is a data representation to formulate the feature representation. In the second phase, model training, testing, and their evaluation is then conducted to evaluate the performance of the proposed method. 

\begin{figure}
\includegraphics[scale=0.7]{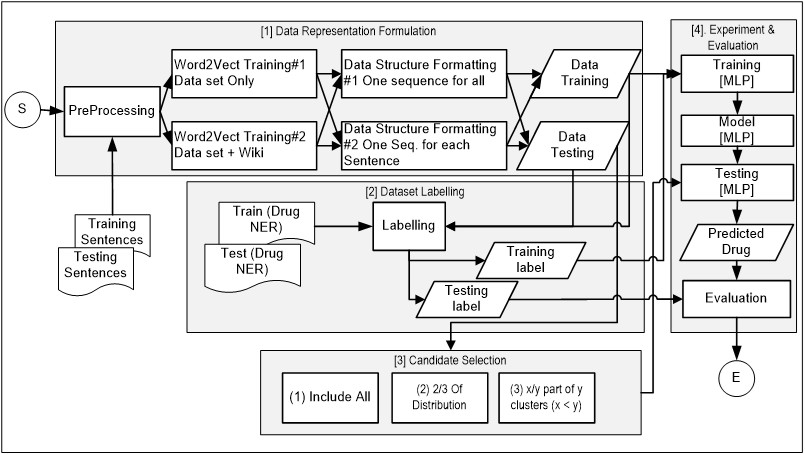}
\caption{Proposed approach framework of the first experiment} 
\label{fig:first_exp}
\end{figure}

The proposed method of the first experiment consists of 4 steps (see Figure \ref{fig:first_exp}). The first step is a data representation formulation. The output of the first step are the tuples of training and testing dataset. The second step is dataset labeling which is applied to both testing and training data. The step provides the label of each tuple. The third step is the candidate selection which is performed to minimize the noises since the actual drug target quantity is a far less compared to nondrug name. In the last step, we performed the experiment with MLP-NN model and its result evaluation. The detailed explanation of each step is explained in the Subsection \ref{data_representation}, \ref{Candidates_Selection}, and \ref{Evaluation}, whereas section \ref{training_data_Analysis} and \ref{Word_Embedding_Analysis} describe training data analysis as the foundation of this proposed method. As a part of the first experiment, we also evaluate the impact of the usage of the Euclidean distance average as the model's regularization. This regularization term is described in the Subsection \ref{MLP}.

\begin{figure}
\includegraphics[scale=0.7]{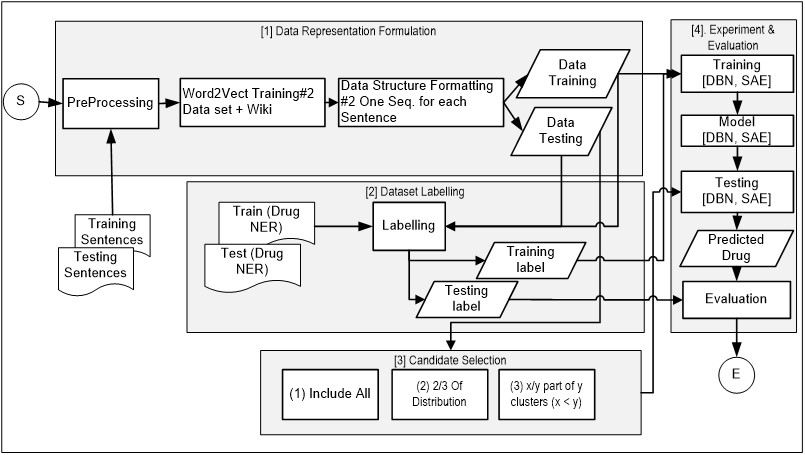}
\caption{Proposed approach framework of the second experiment}
\label{fig:firsta_exp}
\end{figure}

The framework of the second experiment which involves DBN and SAE learning model to the second data representation technique is illustrated in Figure \ref{fig:firsta_exp}. In general, the steps of the second experiment is similar to the first one, with its differences are the data representation used and the learning model involved. In the second experiment, it is used the second technique only with DBN and SAE as the learning model.  

\begin{figure}
\includegraphics[scale=0.7]{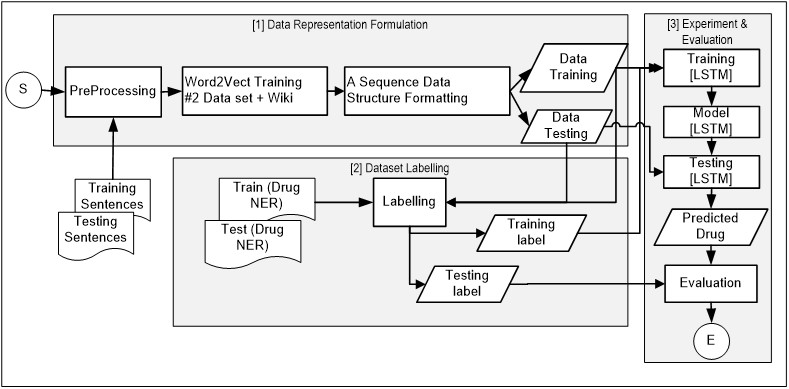}
\caption{Proposed approach framework of the third experiment}
\label{fig:second_exp}
\end{figure}

The framework of the third experiment using the LSTM is illustrated in Figure \ref{fig:second_exp}. There are tree steps in the third experiment. The first step is sequence data representation formulation which provides both sequence training data and testing data. The second step is data labeling which generates the label of training and testing data. LSTM experiment and its result evaluation are performed in the third step. The detail description of these tree step are presented in Subsection \ref{data_representation}, \ref{third_technique}, and  Subsection \ref{Evaluation} as well.

\subsection{Training Data Analysis} 
\label{training_data_Analysis} 
\begin{figure}
\includegraphics[scale=0.9]{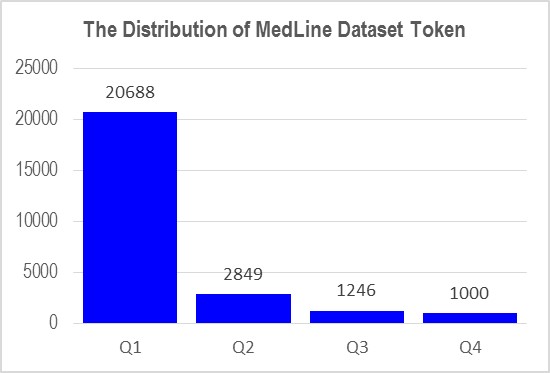}
\caption{Distribution of MedLine train dataset token }
\label{fig:dist_medline}
\end{figure}

Each of the sentences in the dataset contains four data types i.e. drug, group, brand, and drug-n. If the sentence contains none of those four types, the type value is null. In the study, we extracted drug and drug-n. Overall in both DrugBank and MedLine datasets, the quantity of drug name target is far less compared to the non-drug target. Segura et al. \cite{Segura-Bedmar2013} present the first basic statistics of the dataset. A more detailed exploration regarding token distribution in the training dataset  is described in this section. The MedLine sentences training dataset contains 25.783 single token, which consists of 4.003 unique tokens. Those tokens distributions are not uniform, but are dominated by a small part of some unique tokens. If all of the unique tokens are arranged and ranked based on the most frequent appearances in the sentences, the quartile distribution will have the following result presented in Figure  \ref{fig:dist_medline}. Q1 represents the token number 1 to 1001 which their total of frequency is 20.688. Q2 represents the token number 1002 to 2002 which their total of frequency is 2.849. Q3 represents the token number 2003 to 3002 which their total of frequency is 1.264, and Q4 represents the token number 3003 to 4003 which their total of frequency is 1.000. The Figure shows that the majority appearances are dominated by only a small amount of the total tokens. 

\begin{figure}
\includegraphics[scale=0.9]{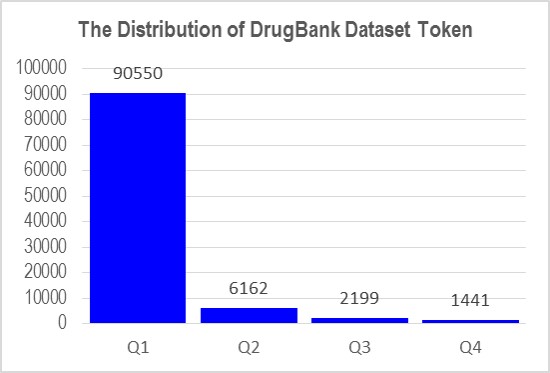}
\caption{Distribution of DrugBank train dataset token }
\label{fig:dist_DrugBank}
\end{figure}

Further analysis of the dataset tokens shows that most of the drug names of the targeted token rarely appear in the dataset. When we divide those token collections into three partitions based on their sum of frequency, as presented in Table \ref{table:MedLine_freqdist}, it is shown that all of the drug name entity targeted are contained in 2/3 part with less frequent appearances of each token (a unique token in the same sum of frequency). The similar pattern of training data token distribution also emerged in the DrugBank dataset as illustrated in Figure \ref{fig:dist_DrugBank} and Table \ref{table:DrugBank_freqdist}. When we look into specific token distributions, the position of most of drug name target are in the third part. Since the most frequently appeared words in the first and the second parts are the most common words such as: stop words ($"$of$"$, $"$the$"$, $"$a$"$, $"$end$"$, $"$to$"$, $"$where$"$, $"$as$"$, $"$from$"$, and such kind of words) and common words in medical domain such as $"$administrator$"$, $"$patient$"$,  $"$effect$"$, $"$dose$"$, etc.  

\begin{table}[h!]
\caption{The frequency distribution $\&$ drug target token position, MedLine }
\label{table:MedLine_freqdist}
\begin{tabular}{|c| r | r | r|} 
\hline
1/3$\#$ &  $\Sigma$ Sample &  $\Sigma$ Frequency  &  $\Sigma$ Single Token of Drug Entity\\ [0.5ex] 
\hline\hline
1&          28 &          8,661 &                    -   \\ 
\hline
2&        410 &          8,510 &                    50 \\
\hline
3&      3,563 &          8,612 &                   262 \\
\hline
\end{tabular}
\end{table}

\begin{table}[h!]
\caption{The frequency distribution $\&$ drug target token position, DrugBank }
\label{table:DrugBank_freqdist}
\begin{tabular}{| r| r | r | r|} 
\hline
1/3$\#$ &  $\Sigma$ Sample &  $\Sigma$ Frequency  &  $\Sigma$ Single Token of Drug Entity\\ [0.5ex] 
\hline\hline
1&          27 &        33,538 &                    -   \\ 
\hline
2&        332 &        33,463 &                    33\\
\hline
3&      5,501 &        33,351 &                  920 \\
\hline
\end{tabular}
\end{table}

\subsection{Word Embedding Analysis}
\label{Word_Embedding_Analysis}
To represent the dataset we utilized the word embedding model proposed by Mikolov et al. \cite{Mikolov2013a}. We treated all of the sentences as a corpus after the training dataset and testing dataset were combined. The used word2vec training model was the CBOW  (Continuous Bag Of Words) model with context window length 5, and the vector dimension 100. The result of the word2vec training is the representation of word in 100 dimension row vectors. Base on the row vector, it can be estimated the similarities or dissimilarities between words. The description below is the analysis summary of word2vec representation result which is used as a base reference for the data representation technique and the experiment scenarios. By taking some sample of drug targets and non-drug vector representation, it is shown that drug word has more similarities (cosine distance) to another drug than to non-drug and the vice versa. Some of those samples are illustrated in Table \ref{table:cosine_dist}. We also computed the Euclidean distance between all of the words. Table \ref{table:avg_eucldist} shows the average of Euclidean distance and cosine distance between drug-drug, drug - non-drug, and non-drug - non-drug. These values of the average distance show us that, intuitively, it is feasible to group the collection of the word into drug group and non-drug group based on their vector representations value.

\begin{table}[h!]
\caption{Some of the cosines-distance similarities between two kinds of words}
\label{table:cosine_dist}
\begin{tabular}{| l  | l | r | l |} 
\hline
\multicolumn{1}{|c|}{\textbf{word 1}}          &	\multicolumn{1}{|c|}{\textbf{word 2}}       &	\multicolumn{1}{|c|}{\textbf{similarities(cosine-dist)}}&	\multicolumn{1}{|c|}{\textbf{remark}}\\
\hline
\hline
dilantin        &	tegretol    &	0.75135758     &	drug-drug\\
\hline
phenytoin       &	dilantin    &	0.62360351     &	drug-drug\\
\hline
phenytoin       &	tegretol    &	0.51322415      &	drug-drug\\
\hline
cholestyramine  &	dilantin    &	0.24557819     &	drug-drug\\
\hline
cholestyramine  &	phenytoin   &	0.23701277     &	drug-drug\\
\hline
administration  &	patients    &	0.20459694     &	non-drug - non-drug\\
\hline
tegretol        &	may         &	0.11605539     &	drug - non-drug\\
\hline
cholestyramine  &	patients    &	0.08827197     &	drug - non-drug\\
\hline
evaluated       &  	end         &	0.07379115     &	non-drug - non-drug\\
\hline
within          &	controlled  & 	0.06111103     &	non-drug - non-drug\\
\hline
cholestyramine  &	evaluated   &	0.04024139     &	drug - non-drug\\
\hline
dilantin        &	end         &	0.02234770     &	drug - non-drug\\
\hline
\end{tabular}
\end{table}

\begin{table}[h!]
\caption{The average of euclidean-distance and cosine similarities between groups of word}
\label{table:avg_eucldist}
\begin{tabular}{| l | r | r |} 
\hline
\multicolumn{1}{|c|}{\textbf{Word Group}}      &   \multicolumn{1}{|c|}{\textbf{Euclidean Dist. avg}} & \multicolumn{1}{|c|}{\textbf{Cosine Dist. avg}} \\ 
\hline\hline
drug - non-drug    &   0.096113798 &0.194855980\\
\hline
non-drug - non-drug &   0.094824332 &0.604091044\\
\hline
drug-drug       &   0.093840800   &0.617929002\\
\hline
\end{tabular}
\end{table}

\subsection{Feature Representation, Data Formatting, $\&$ Data Labelling}
\label{data_representation}
Based on the training data and word embedding analysis, we formulate the feature representation and its data formatting. In the first and second techniques, we try to overcome the multiple tokens drawback that left unsolved in \cite{Bjorne2013} by formatting a single input data as an N - gram model with N=5 (one tuple data consist 5 tokens) to accommodate the maximum token which acts as a single drug entity target name. The tuples were provided from the sentences of both training and testing data. Thus, we have a set of tuples of training data and a set of tuples of testing data. Each tuple was treated as a single input.

\begin{algorithm}
\smaller
\SetAlgoLined
\KwResult{Labelled dataset }
\textbf{Input:}array of tuple, array of drug \;
\textbf{output}: array of label \{Array of drug contains list of drug and drug-n only\} \;
 label[]$<$$=$1 {Initialization}\;
 \For{each t in tuple}{
   \For{each d in drug}{
        \eIf{length (d) = 1}{
            \eIf{t[1] = d[1]} 
            {//match 1 token drug\;
             label $<$$=$ 2, break, {exit from for each d in drug}\;}
            {}
        }{
          \eIf{length (d) = 2} {
            \eIf{t[1] = d[1] and t[2] = d[2]} 
            {//match 2 tokens drug\;
             label $<$$=$ 3, break, {exit from for each d in drug}\;}
            {}
          }
          {
            \eIf{length (d) = 3}
                {
                \eIf{t[1] = d[1] and t[2] = d[2] and t[3] = d[3]} 
                {
                 label $<$$=$ 4,
                 break, {exit from for each d in drug}\;}
                {}
            }
            {
                \eIf{length (d) = 4}{
                \eIf{t[1] = d[1] and t[2] = d[2] and t[3] = d[3] and t[4] = d[4]}
                {
                 label $<$$=$ 5,
                 break, {exit from for each d in drug}\;
                }
                {}
                }
                {
                \eIf{length (d) = 5}{
                \eIf{t[1] = d[1] and t[2] = d[2] and t[3] = d[3] and t[4] = d[4] and t[5] = d[5]  }{
                 label $<$$=$ 6,
                 break, {exit from for each d in drug}\;
                }
                {}
                }
                {}
                }
            }
          } 
        }
      } 
  }
  \caption{Dataset Labelling}
  \label{algo_label}
\end{algorithm}

To identify a single input, whether it is a non-drug or drug target, we use a multi-classification approach which classifies the single input into one of six classes. class 1 represents non-drug whereas the other classes represent drug target which also identified how many tokens (word) that perform the drug target. To identify which class a certain tuple belongs to is determined as follows: The drug tuple is the tuple which its first token (token-1) is the drug type. If the token-1 is not a drug, regardless of whatever the rest of the 4 tokens are, then the tuple is classified as no drug. This kind of tuple is identified as class 1. If the token-1 is a drug, and token-2 is not a drug, regardless of the last 3 tokens, the tuple will be identified as class 2, and so on. 

Since we only extracted the drug entity, we ignored the other token types, whether it is a group, brand, or another common token. To provide the label of each tuple, we only use the drug and drug-n types as the tuple reference list. In general, if the sequence of token in each tuple in dataset contains the sequence which is exactly same with one of tuple reference list members, then the tuple in dataset is identified as drug entity. The detail of the algorithm used to provide the label of each tuple in both training data and testing data is described in Algorithm \ref{algo_label}:

We proposed two techniques in constructing the tuple set of the sentences. The first technique treats all sentences as one sequence, whereas in the second technique, each sentence is processed as one sequence. The first and the second techniques are evaluated with MLP, DBN, and SAE model. The third technique treats the sentences of dataset as a sequence where the occurrence of the current token is influenced by the previous one. By treating the sentence as a sequence not only in the data representation but also in the classification and recognition process, the most suitable model to be used is RNN. We applied RNN-LSTM to the third technique.

\subsubsection{First Technique}
\label{first_technique}
The first dataset formatting (one sequence for all sentences) is performed as follows. In the first step, all sentences in the dataset are formatted as a token sequence. Let the token sequence is: \[t_1  t_2 t_3 t_4 t_5 t_6 t_7 t_8...t_n\] with $n$ is number of token in the sequences,  then the dataset format will be: \[t_1 t_2 t_3 t_4 t_5;t_2 t_3 t_4 t_5 t_6;…..t_{n-4} t_{n-3} t_{n-2} t_{n-1} t_n;\] 

A sample of sentences and their drug name are presented in Table \ref{table:DrugBank_sentence}.Taken from DrugBank training data Table \ref{table:DrugBank_sentence} is the raw data of 3 samples with three relevant fields, i.e., sentences, character drug position, and the drug name. Table \ref{table:datasetform_1} illustrates a portion of the dataset and its label as the result of the raw data in Table \ref{table:DrugBank_sentence}. Refer to the drug-n name field in the dataset, dataset number 6 is identified as a drug, whereas the others are classified as a non-drug entity. The complete label illustration of the dataset provided by the first technique is presented in Table \ref{table:first_tech}. As described in the Section \ref{data_representation}, the value of vector dimension for each token is 100. Therefore, for a single data, it is represented as 100*5 = 500 lengths of a one-dimensional vector.

\begin{table}[h!]
\small
\caption{Sample of DrugBank sentences and their drug name target}
\label{table:DrugBank_sentence}
\begin{tabular}{ |p{5cm}|p{1.5cm}|p{2.0cm}|  }
\hline
\multicolumn{1}{|c|}{\textbf{Sentence}}&\multicolumn{1}{|c|}{\textbf{Drug}}&\multicolumn{1}{|c|}{\textbf{Drug}}\\
\multicolumn{1}{|c|}{}&\multicolumn{1}{|c|}{\textbf{Position}}&\multicolumn{1}{|c|}{\textbf{Name}}\\
\hline\hline
modification of surface histidine&&clostridium\\
residues abolishes the cytotoxic&79-107&difficile\\
activity of clostridium difficile&&toxin a\\
toxin a&&\\
\hline
antimicrobial activity of ganoderma&&ganoderma\\
lucidum extract alone and in&26-50&lucidum\\
with some antibiotics.&&extract\\
\hline
on the other hand, surprisingly,&&green tea\\
green tea gallocatechins,&&gallocatechins\\
(-)-epigallocatechin-3-o-gallate and&33-56&\\
theasinensin a, potently enhanced&&\\
the promoter activity (182 and 247$\%$&&\\
activity at 1 microm, respectively).&&\\
\hline
\end{tabular}
\end{table}

\begin{table}[h!]
\small
\caption{A portion of the dataset formulation as the results of DrugBank sample with first technique}
\label{table:datasetform_1}
\begin{tabular}{| p{1cm} | p{2cm} | p{1cm} | p{1cm} |p{2cm} | p{2cm} | p{0.5cm}|} 
\hline
\textbf{dataset Number}&\multicolumn{1}{|c|}{\textbf{token-1}}&\multicolumn{1}{|c|}{\textbf{token-2}}
&\multicolumn{1}{|c|}{\textbf{token-3}}&\multicolumn{1}{|c|}{\textbf{token-4}}&\multicolumn{1}{|c|}{\textbf{token-5}}&\multicolumn{1}{|c|}{\textbf{label}}
\\
\hline\hline
1&modification&of&surface&histidine&residues&1\\
\hline
2&of&surface&histidine&residues&abolishes&1\\
\hline
3&surface&histidine&residues&abolishes&the&1\\
\hline
4&histidine&residues&abolishes&the&cytotoxic&1\\
\hline
5&the&cytotoxic&activity&of&clostridium&1\\
\hline
6&\textbf{clostridium} &\textbf{difficile} &\textbf{toxin} &\textbf{a}&antimicrobial&5\\
\hline
7&difficile &toxin &a&antimicrobial&activity&1\\
\hline
\end{tabular}
\end{table}

\begin{table}[h!]
\small
\caption{First Technique Data Representation and Its Label}
\label{table:first_tech}
\begin{tabular}{| p{2cm} | p{2.5cm} | p{2cm} |p{2cm} | p{1.5cm} | p{0.5cm}|} 
\hline
\multicolumn{1}{|c|}{\textbf{token-1}}&\multicolumn{1}{|c|}{\textbf{token-2}}&\multicolumn{1}{|c|}{\textbf{token-3}}&\multicolumn{1}{|c|}{\textbf{token-4}}&\multicolumn{1}{|c|}{\textbf{token-5}}
&\multicolumn{1}{|c|}{\textbf{label}}\\
\hline
\hline
\textbf{'plenaxis'}	&'were'	&'performed'	&'cytochrome'	&'p-450'	&2\\
\hline
\textbf{'testosterone'}	&'concentrations'	&'just'	&'prior'	&'to'	&2\\
\hline
\textbf{'beta-adrenergic'}	&\textbf{'antagonists'}	&'and'	&'alpha-adrenergic'	&'stimulants,'	&3\\
\hline
\textbf{'carbonic'}	&\textbf{'anhydrase'}	&'inhibitors,'	&'concomitant'	&'use'	&3\\
\hline
\textbf{'sodium'}	&\textbf{'polystyrene'}	&\textbf{'sulfonate'}	&'should'	&'be'	&4\\
\hline
\textbf{'sodium'}	&\textbf{'acid'}	&\textbf{'phosphate'}	&'such'	&'as'	&4\\
\hline
\textbf{'clostridium'}	&\textbf{'difficile'}	&\textbf{toxin'}	&\textbf{a'}	&'-'	&5\\
\hline
\textbf{'nonsteroidal'}	&\textbf{anti'}	&\textbf{'inflammatory'}	&\textbf{'drugs'}	&'and'	&5\\
\hline
\textbf{'casein'}	&\textbf{'phosphopeptide-amorphous'}
&\textbf{'calcium'}	&\textbf{'phosphate'}	&\textbf{'complex'}	&6\\
\hline
'studies'	&'with'	&'plenaxis'	&'were'	&'performed.'	&1\\
\hline
'were'	&'performed.'	&'cytochrome'	&'p-450'	&'is'	&1\\
\hline
\end{tabular}
\end{table}

\subsubsection{Second Technique}
\label{second_technique}
The second technique is used for treating one sequence that comes from each sentence of the dataset. With this treatment, we added special characters $*$, as a padding, to the last part of the token when its dataset length is less than 5. By applying the second technique the first sentence of the sample provided a dataset as illustrated in Table \ref{table:second_tech}.

\begin{table}[h!]
\small
\caption{Second technique data representation and its label}
\label{table:second_tech}
\begin{tabular}{| p{2cm} | p{1.5cm} | p{1.5cm} |p{1.5cm} | p{1.5cm} | p{1cm}|} 
\hline
\multicolumn{1}{|c|}{\textbf{token-1}}&\multicolumn{1}{|c|}{\textbf{token-2}}&\multicolumn{1}{|c|}{\textbf{token-3}}&\multicolumn{1}{|c|}{\textbf{token-4}}&\multicolumn{1}{|c|}{\textbf{token-5}}
&\multicolumn{1}{|c|}{\textbf{label}}\\
\hline
'modification'&'of'&'surface'&'histidine'&'residues'&1\\
\hline
'of'&'surface'&'histidine'&'residues'&'abolishes'&1\\
\hline
surface&histidine&residues&abolishes&the&1\\
\hline
'histidine'&'residues'&'abolishes'&'the'&'cytotoxic'&1\\
\hline
'the'&'cytotoxic'&'activity'&'of'&'clostridium'&1\\
\hline
\textbf{'clostridium'}&\textbf{'difficile'}&\textbf{'toxin'}&'a'&'*'&5\\
\hline
'difficile'&'toxin'&'a'&'*'&'*'&1\\
\hline
'a'&'atoxin'&'*'&'*'&'*'&1\\
\hline
'toxic'&'*'&'*'&'*'&'*'&1\\
\hline
\end{tabular}
\end{table}

\subsubsection{Third Technique}
\label{third_technique}
Naturally, the NLP sentence is a sequence in which the occurrence of the current word is conditioned by the previous one. Based on the word2vec value analysis, it is shown that intuitively we can separate the drug word and non-drug word by their Euclidean distance. Therefore, we used the Euclidean distance between the current words with the previous one to represent the influence. Thus, each current input $x_i$ is represented by $[xv_i  xd_i]$ which is the concatenation of word2vec value $xv_i$ and its Euclidian distance to the previous one, $xd_i$. Each x is the row vector with the dimension length is 200, the first 100 values are its word2vector, and the rest of all 100 values are the Euclidian distance to the previous. For the first word all value of $xd_i$ is 0. With the LSTM model, the task to extract the drug name from the medical data text is the binary classification that applied to each word of the sentence. We formulate the word sequence and its class as described in Table \ref{table:third_tech}. In this experiment, each word that represents the drug name is identified as class 1, such as 'plenaxis', 'cytochrome', and 'p-450,' whereas the other words are identified by class 0.

\begin{table}[h!]
\small
\caption{Third technique of data representation and its label}
\label{table:third_tech}
\hspace{-1.7cm}
\begin{tabular}{ |l|c|c|c|c|c|c|c|c|  }
\hline
 \multicolumn {1}{|c|}{\multirow{2}{*}{Sent.\#1}} & Class&0&0&0&0&1&0&0\\
 \cline {2-9}
 \multicolumn {1}{|c|}{\multirow{2}{*}{}}&Word&'drug'&'interaction'&'studies'&'with'&'plenaxis'&'were'&'performed'\\
 \hline
 \hline
 \multicolumn {1}{|c|}{\multirow{2}{*}{Sent.\#2}} & Class&1&1&0&0&0&0&0\\
 \cline {2-9}
 \multicolumn {1}{|c|}{\multirow{2}{*}{}}&Word&'cytochrome'&'p-450'&'is'&'not'&'known'&in'&'the\\
  \hline
\end{tabular}
 \end{table}

\subsection{Wiki Sources}
\label{Wiki}
 In this study we also utilize Wikipedia as the additional text sources in word2vec training as used by \cite{Segura-bedmar2015}. The Wiki text addition is used to evaluate the impact of the training data volume in improving the quality of word's vector.

\subsection{Candidates Selection}
\label{Candidates_Selection}
The tokens as the drug entities target are only a tiny part of the total tokens. In MedLine dataset, 171 of 2.000 token (less than ten \%)  are drugs, whereas in DrugBank, the number of drug tokens are 180 of 5.252 \cite{Segura-Bedmar2013}. So the major part of these tokens are non-drug and other noises such as a stop word, and special or numerical characters. Based on this fact, we propose a candidate selection step to eliminate those noises. We examine two mechanisms in the candidate selection. The first is based on token distribution. The second is formed by selecting $x/y$ part of the clustering result of data test. In the first scenario, we only used 2/3 of the token, which appears in the lower 2/3 part of the total token. This is presented in Table \ref{table:MedLine_freqdist} and Table \ref{table:DrugBank_freqdist}. Whereas, in the second mechanism we selected $x/y (x < y)$ which is a part of total token after the tokens are clustered into y clusters.

\subsection{Overview of NN Model}
\subsubsection{MLP}
\label{MLP}
In the first experiment, we used multi-layer perceptron NN to train the model and evaluate the performance \cite{LeCun2015}. Given a training set of $m$ examples, then the overall cost function can be defined as:

\begin{equation}
    J(W,b)=\Bigg[\frac{1}{m} \sum_{i=1}^{m} J(W,b;x^i,y^i)\Bigg] + \frac{\lambda}{2} \sum_{l=1}^{nl-1} \sum_{i=1}^{sl} \sum_{j=1}^{sl-1} \Big(W_{ji}^{(l)} \Big)^2 
\end{equation}
\begin{equation}
    J(W,b)=\Bigg[\frac{1}{m} \sum_{i=1}^{m} \Bigg(\frac{1}{2}\Bigg|\Bigg|h_{wb}\Big(x^{(i)}\Big)-y^i\Bigg|\Bigg|^2  \Bigg)\Bigg]+\frac{\lambda}{2}\sum_{l=1}^{nl-1}\sum_{i=1}^{sl} \sum_{j=1}^{sl-1}\Big(W_{ji}^{(l)} \Big)^2 
\end{equation}

In the definition of $J(W,b)$, the first term is an average sum-of-squares error term, whereas the second term is a regularization term which is also called a weight decay term. In this experiment we use three kinds of regularization:
$\#$0, L0 with $\lambda$ = 0,
$\#$1, L1 with $\lambda$ = 1, and $\#2$ with $\lambda$ = the average of Euclidean distance. We computed the L2's $\lambda$ based on the word embedding vector analysis that drug target and non-drug can be distinguished by looking at their Euclidean distance. Thus, for L2 regularization, the parameter is calculated as: 
\begin{equation}
    \lambda=\frac{1}{n*(n-1)} \sum_{i=1}^{n-1} \sum_{j=i+1}^{n} \textit{dist}(x^i,x^j)
\end{equation}
\noindent where \textit{dist}($x^i$,$x^j$) is the Euclidean distance of $x^i$ and $x^j$.

The model training and testing are implemented by modifying the code from \cite{IMM2012-06284} which can be downloaded at  https://github.com/rasmusbergpalm/DeepLearnToolbox. 



\subsubsection{DBN}
\label{DBN}
DBN is a learning model composed of two or more stacked RBM \cite{Hinton2006,Fischer2012}. An RBM is an undirected graph learning model which associates with a Markov Random Fields (MRF). In the DBN, the RBM acts as feature extractor where the pre-training process provides initial weights values to be fine-tuned in the discriminative process in the last layer. The last layer may be formed by logistic regression or any standard discriminative classifiers \cite{Fischer2012}. RBM was originally developed for binary data observation \cite{Tieleman2008,Dahl2012}. It is a popular type of unsupervised model for binary data \cite{Tang2011,Hinton2010}. Some derivative of RBM models are also proposed to tackle a continuous/real values suggested in \cite{H.C2003,Welling2005}. 

\subsubsection{SAE}
\label{SAE}
An autoencoder (AE) neural network is one of the unsupervised learning algorithms. The NN tries to learn a function $h(w,x) \approx x$. The autoencoder NN architecture also consists of input, hidden, and output layers. The particular characteristic of the autoencoder is that the target output is similar to the input. The interesting structure of the data is estimated by applying a certain constraint to the network, which limits the number of hidden units. However, when the number of hidden units has to be larger, it can be imposed with sparsity constraints on the hidden units \cite{Ng}. The sparsity constraint is used to enforce the average value of hidden unit activation constrained to a certain value. As used in the DBN model, after we trained the SAE, the trained weight was used to initialize the weight of NN for the classification.

\subsubsection{RNN-LSTM}
\label{RNN_LSTM}
RNN (Recurrent Neural Network) is an NN, which considers the previous input in determining the output of the current input. RNN is powerful when it is applied to the dataset with a sequential pattern or when the current state input depends on the previous one, such as the time series data, sentences of NLP. An LSTM network is special kind of RNN which also consists of 3 layers, i.e., an input layer, a single recurrent hidden layer, and an output layer \cite{Hochreiter1997}. The main innovation of LSTM is that its hidden layer consists of one or more memory blocks. Each block includes one or more memory cells. In the standard form, the inputs are connected to all of the cells and gates, whereas the cells are connected to the outputs. The gates are connected to other gates and cells in the hidden layer. The single standard LSTM is a hidden layer with input, memory cell, and output gates \cite{Hammerton2003,Olah2015}. 

\subsection{Dataset}
\label{Dataset}
To validate the proposed approach, we utilized DrugBank and MedLine open dataset, which have also been used by previous researchers. Additionally, we used drug label documents from various drug producers and regulator Internet sites located in Indonesia:

\begin{enumerate}
    \item http://www.kalbemed.com/
    \item http://www.dechacare.com/
    \item http://infoobatindonesia.com/obat/, and
    \item http://www.pom.go.id/webreg/index.php/home/produk/01.
\end{enumerate}
  
The drug labels are written in Bahasa Indonesia, and their common contents are drug name, drug components, indication, contra indication, dosage, and warning. 

\subsection{Evaluation}
\label{Evaluation}
To evaluate the performance of the proposed method, we use common measured parameters in data mining i.e., precision, recall, and F-score. The computation formula of these parameters is as follows. Let $C =  \{C_1, C_2, C_3 ...C_n\}$ is a set of the extracted drug-name of this method, and $K= \{K_1, K_2, K_3,...K_l\}$ is set of actual drug-name in the document set D.  Adopted from \cite{Sadikin2014}, the parameter computations formula are:


\begin{equation}
Precision (K_i,C_j)=\frac{(True Positive)}{(True Positive+False Positive )}=\frac{(||K_i \cap C_j||)}{(||C_j||)}
\end{equation}

\begin{equation}
Recall (K_i,C_j)=\frac{(True Positive)}{(True Positive+False Negative )}=\frac{(||K_i \cap C_j||)}{(||K_i||)} 
\end{equation}

\noindent where $\|K_i\|, \|C_j\|$, and $\|K_i \cap C_j\|$ denote the number of drug-name in $K$, in $C$, and in both $K$ and $C$ respectively. The F-score value is computed by the below formula:

\begin{equation}
F-score (K_i,C_j)=\frac{(2*Precision (K_i,C_j)*Recall (K_i,C_j))}{(True Positive+False Positive )}
\end{equation}

\section{Results and Discussion}
\label{result_and_discussion}
\subsection{MLP Learning Performance}
\label{MLP_Learning_Performance}

\begin{figure}
\includegraphics[scale=0.55]{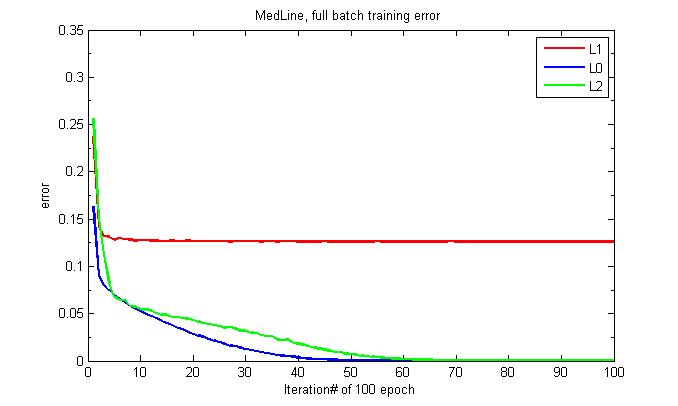}
\caption{Full batch training error of MedLine dataset }
\label{fig:MedLine_regular}
\end{figure}

\begin{figure}
\includegraphics[scale=0.55]{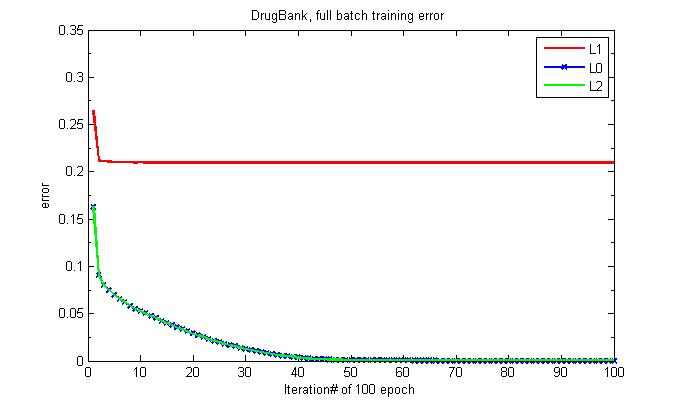}
\caption{Full batch training error of DrugBank dataset }
\label{fig:DrugBank_regular}
\end{figure}

The following experiments are the part of the first experiment. These experiments are performed to evaluate the contribution of the three regularization settings as described in the Subsection \ref{MLP} . By arranging the sentence in training dataset as 5-gram of words, the quantity of generated sample is presented in Table \ref{table:dataset_comp}. We do training and testing of the MLP-NN learning model for all those test data compositions. The result of model performances on both datasets, i.e., MedLine \& DrugBank, in learning phase is shown Figure \ref{fig:MedLine_regular} and Figure \ref{fig:DrugBank_regular}. The NN Learning parameters that are used for all experiments are: 500 input nodes, two hidden layers where each layer has 100 nodes with sigmoid activation, and 6 output nodes with softmax function; the learning rate = 1, momentum = 0.5; and epochs = 100. We used mini-batch scenario in the training with the batch size is 100. The presented errors in Figure \ref{fig:MedLine_regular} and Figure \ref{fig:DrugBank_regular} are the errors for full batch, i.e., the mean errors of all mini batches.

The learning model performance shows different patterns between MedLine and DrugBank datasets. For both datasets, L1 regularization tends to stabilize in the lower iteration and its training error performance is always less than L0 or L2. The L0 and L2 training error performance pattern, however, shows a slight different behavior between MedLine and DrugBank. For the MedLine dataset, L0 and L2 produce different results for some of iterations. Whereas, the training error performance of L0 and L2 for DrugBank are almost the same in every iteration. Different pattern results are probably due to the variation in the quantity of training data. As illustrated in Table \ref{table:dataset_comp}, the volume of DrugBank training data is almost four times the volume of the MedLine dataset. It can be concluded that, for larger dataset, the contribution of L2 regularization setting is not too significant in achieving better performance. For smaller dataset (MedLine), however, the performance is better even after only few iterations.   

\begin{table}[h!]
\caption{Dataset Composition}
\label{table:dataset_comp}
\begin{tabular}{ |l|r|r|r|r|  }
\hline
 \multicolumn {1}{|c|}{\multirow{2}{*}{\textbf{dataset}}} &  \multicolumn {1}{|c|}{\multirow{2}{*}{\textbf{Train}}} &\multicolumn{3}{|c|}{\textbf{Test}} \\
 \cline {3-5}
 
 \multirow{2}{4em}{}& \multirow{2}{4em}{}& \textbf{All} 	& \textbf{2/3 Part} &\textbf{Cluster}\\
 \hline
    MedLine&        26,500 &      10,360 &      6,673 &      5,783 \\
    DrugBank&      100,100 &        2,000 &      1,326 &      1,933 \\
  \hline
 \end{tabular}
 \end{table}
 
\subsection{Open Dataset Performance}
\label{Open_Dataset_Performance}
 In the Table \ref{table:fscore_1}, \ref{table:fscore_2}, \ref{table:fscore_3}, and \ref{table:fscore_MLP_DBN_SAE}, the numbering (1), (2) and (3) in the most left column indicate the candidate selection technique with:
\begin{enumerate}
    \item (1) : all data test are selected
    \item (2) : 2/3 part of data test are selected, and
    \item (3) : 2/3 part of 3 clusters for MedLine or 3/4 part of 4 cluster for DrugBank 
\end{enumerate}

\subsubsection{MLP-NN Performance }
\label{MLP_NN_Performance}

In this first experiment, for two data representation techniques and three candidate selection scenarios, we have six experiment scenarios. The result of the experiment which applies the first data representation technique and three candidate selection scenarios is presented in Table \ref{table:fscore_1}. In computing the F-score, we only select the predicted target which is provided by the lowest error (the minimum one). For MedLine dataset, the best performance is shown by L2 regularization setting where the error is 0.041818, in third candidate selection scenario with F-score 0.439516, whereas the DrugBank is achieved together by L0 and L1 regularization setting, with an error test of 0.0802, in second candidate selection scenario, the F-score was 0.641745. Overall, it can be concluded that DrugBank experiments give the best F-score performance. The candidate selection scenarios also contributed to improving the performance, as we found for both of MedLine and DrugBank, the best achievement is provided by the second and third scenarios respectively. 

\begin{table}[h!]
\centering
\caption{The F-score performances of three of scenarios experiments }
\label{table:fscore_1}
\begin{tabular}{| l | r | r | r | r | l |} 
\hline
\multicolumn{1}{|c|}{\textbf{MedLine}}	&\multicolumn{1}{|c|}{\textbf{Prec}}	&\multicolumn{1}{|c|}{\textbf{Rec}} &\multicolumn{1}{|c|}{\textbf{F-score}} 
&\multicolumn{1}{|c|}{\textbf{Lx}}
&\multicolumn{1}{|c|}{\textbf{error Test}} \\
\hline
\hline
(1)&0.3564&0.5450	&0.4310	&L0	&0.0305\\
\hline
(2)&0.3806	&0.5023	&0.4331	&L1,L2	&0.0432\\
\hline
(3)&0.3773	&0.5266	&0.4395	&L2	&0.0418\\
\hline
\hline
\multicolumn{1}{|c|}{\textbf{DrugBank}}	&\multicolumn{1}{|c|}{\textbf{Prec}}	&\multicolumn{1}{|c|}{\textbf{Rec}} &\multicolumn{1}{|c|}{\textbf{F-score}} 
&\multicolumn{1}{|c|}{\textbf{Lx}}
&\multicolumn{1}{|c|}{\textbf{error Test}} \\
\hline
\hline
(1)&0.6312	&0.5372	&0.5805	&L0	&0.07900\\
\hline
(2)&0.6438	&0.6398	&0.6417	&L0,L2	&0.0802\\
\hline
(3)	&0.6305	&0.5380	&0.5806	&L0	&0.0776\\
\hline
\end{tabular}
\end{table}
The next experimental scenario in the first experiment is performed to evaluate the impact of the data representation technique and the addition of Wiki source in word2vec training. The results are presented in Table \ref{table:fscore_2} and Table \ref{table:fscore_3}. According to the obtained results presented in Table \ref{table:fscore_1}, the L0 regularization gives the best F-score. Hence, accordingly we only used the L0 regularization for the next experimental scenario. The table \ref{table:fscore_2} presents the impact of the data representation technique. Looking at the F-score, the second technique gives better results for both datasets, i.e., the MedLine and DrugBank.   

Table \ref{table:fscore_3} shows the result of adding the Wiki source into word2vec training in providing the vector of word representation. These results confirm that the addition of training data will improve the performance. It might be due to the fact that most of the targeted token such as drug name, are uncommon words, whereas the words that are used in Wiki's sentence are commonly used words. Hence, the addition of commonly used words will make the difference between drug token and the non-drug token (the commonly used token) becomes greater. For the MLP-NN experimental results, the $4^{th}$ scenario, i.e the second data representation with 2/3 partition data selection in Drugbank dataset, provides the best performance with 0.684646757 in F-score. 

\begin{table}[h!]
\centering
\caption{The F-score performance as an impact of data representation technique }
\label{table:fscore_2}
\begin{tabular}{ |l|r|r|r|r|r|r|  }
\hline
  \multicolumn {1}{|c|}{\textbf{dataset}} &\multicolumn{3}{|c|}{\textbf{(1)One Seq. of all sentences}}
  &\multicolumn{3}{|c|}{\textbf{(2)One Seq. of each Sentence }} \\
\hline
\hline
\multicolumn{1}{|c|}{\textbf{MedLine}}	&\multicolumn{1}{|c|}{\textbf{Prec}}	&\multicolumn{1}{|c|}{\textbf{Rec}} &\multicolumn{1}{|c|}{\textbf{F-score}} 
&\multicolumn{1}{|c|}{\textbf{Prec}}	&\multicolumn{1}{|c|}{\textbf{Rec}} &\multicolumn{1}{|c|}{\textbf{F-score}}\\ 
\hline
(1)& 0.3564& 0.5450& 0.4310& 0.6515& 0.6220& 0.6364\\
(2)& 0.3806& 0.5023& 0.4331& 0.6119& 0.7377& \textbf{0.6689}\\
(3)& 0.3772& 0.5266& \textbf{0.4395}& 0.6143& 0.656873& 0.6348\\
\hline
\hline
\multicolumn{1}{|c|}{\textbf{DrugBank}}	& \multicolumn{1}{|c|}{\textbf{Prec}}	& \multicolumn{1}{|c|}{\textbf{Rec}} & \multicolumn{1}{|c|}{\textbf{F-score}} 
& \multicolumn{1}{|c|}{\textbf{Prec}}	& \multicolumn{1}{|c|}{\textbf{Rec}} & \multicolumn{1}{|c|}{\textbf{F-score}}\\ 
\hline
(1)&0.6438& 0.5337& 0.5836& 0.7143& 0.4962& 0.5856\\
(2)& 0.6438& 0.6398& \textbf{0.6418}& 0.7182& 0.5804& \textbf{0.6420}\\
(3)& 0.6306& 0.5380& 0.5807& 0.5974& 0.5476& 0.5714\\
\hline
\hline
\end{tabular}
 \end{table}


\begin{table}[h!]
\centering
\caption{The F-score performances as an impact of the Wiki addition of word2vec training data}
\label{table:fscore_3}
\begin{tabular}{ |l|r|r|r|r|r|r|  }
\hline
  \multicolumn {1}{|c|}{\textbf{dataset}} &\multicolumn{3}{|c|}{\textbf{(1)One Seq. of all sentences}}
  &\multicolumn{3}{|c|}{\textbf{(2)One Seq. of each Sentence }} \\
\hline
\hline
\multicolumn{1}{|c|}{\textbf{MedLine}}	&\multicolumn{1}{|c|}{\textbf{Prec}}	&\multicolumn{1}{|c|}{\textbf{Rec}} &\multicolumn{1}{|c|}{\textbf{F-score}} 
&\multicolumn{1}{|c|}{\textbf{Prec}}	&\multicolumn{1}{|c|}{\textbf{Rec}} &\multicolumn{1}{|c|}{\textbf{F-score}}\\ 
\hline
(1)& 0.5661 & 0.4582 & 0.5065 & 0.614 & 0.6495 & 0.6336\\
(2)& 0.5661 & 0.4946 & \textbf{0.5279} & 0.5972 & 0.7454 & 0.6631\\
(3)& 0.5714 & 0.4462 & 0.5011 & 0.6193 & 0.6927 & \textbf{0.6540}\\
\hline
\hline
\multicolumn{1}{|c|}{\textbf{DrugBank}}	& \multicolumn{1}{|c|}{\textbf{Prec}}	& \multicolumn{1}{|c|}{\textbf{Rec}} & \multicolumn{1}{|c|}{\textbf{F-score}} 
& \multicolumn{1}{|c|}{\textbf{Prec}}	& \multicolumn{1}{|c|}{\textbf{Rec}} & \multicolumn{1}{|c|}{\textbf{F-score}}\\ 
\hline
(1)& 0.6778 & 0.5460 & 0.6047 & 0.6973 & 0.6107 & 0.6511\\
(2)& 0.6776 & 0.6124 & \textbf{0.6433} & 0.6961 & 0.6736 & \textbf{0.6846}\\
(3)& 0.7173 & 0.5574 & 0.6273 & 0.6976 & 0.6193 & 0.6561\\
\hline
\hline
\end{tabular}
 \end{table}

\subsubsection{DBN $\&$ SAE Performance}
\label{DBN_SAE_Performance}
In the second experiment, which involves DBN and SAE learning model, we only use the experiment scenario that gives the best results in the first experiment. The best experiment scenario uses the second data representation technique with Wiki text as an additional source in the word2vec training step. 

In the DBN experiment, we use two stacked RBMs with 500 nodes of visible unit, 100 nodes of the hidden layer for the first and also the second RBMs. The used learning parameters are as follows: momentum = 0; and alpha= 1. We used mini-batch scenario in the training, with the batch size of 100.  As for RBM constraints, the range of input data value is restricted to $[0..1]$ - as the original RBM, which is developed for binary data type -, whereas the range of vector of word value is $[-1..1]$. So we normalize the data value into $[0..1]$ range before performing the RBM training. In the last layer of DBN, we use one layer of MLP with 100 hidden nodes and 6 output nodes with softmax output function as classifier.   

The used SAE architecture is two stacked AEs with the following nodes configuration. The first AE has 500 units of visible unit, 100 hidden layers, 500 output layer. The second AE has 100 nodes of visible unit, 100 nodes hidden unit, and 100 nodes output unit. The used learning parameters for first SAE and the second SAE respectively are as follows: activation function = sigmoid and tanh; learning rate = 1 and 2; momentum = 0.5 and 0.5; sparsity target = 0.05 and 0.05. The batch size of 100 is set for both of AEs. In the SAE experiment, we use the same discriminative layer as DBN i.e. one layer MLP with 100 hidden node and 6 output nodes with softmax activation function.   

The experiments results are presented in Table \ref{table:fscore_MLP_DBN_SAE}. There is a difference in performances when using the MedLine and the DrugBank datasets when feeding them into MLP, DBN, and SAE models. The best results for the MedLine dataset is obtained when using the SAE. For the DrugBank, the MLP gives the best results. The DBN gives lower average performance for both datasets. The lower performance is probably due to the normalization on the word vector value to $[0...1]$, whereas their original value range is in fact between $[-1..1]$. The best performance for all experiments 1 and 2, is given by SAE, with the second scenario of candidate selection as described in the Subsection \ref{Candidates_Selection}. Its F-score is 0.686192469.

\begin{table}[h!]
\small
\caption{Experimental results of three NN-Model}
\label{table:fscore_MLP_DBN_SAE}
\hspace{-1.5cm}
\begin{tabular}{ |l|r|r|r|r|r|r|r|r|r|}
\hline
  \multicolumn {1}{|c|}{\textbf{dataset}} &\multicolumn{3}{|c|}{\textbf{MLP}}
  &\multicolumn{3}{|c|}{\textbf{DBN}}
  &\multicolumn{3}{|c|}{\textbf{SAE}}\\
\hline
\hline
\multicolumn{1}{|c|}{\textbf{MedLine}}	&\multicolumn{1}{|c|}{\textbf{Prec}}	&\multicolumn{1}{|c|}{\textbf{Rec}} &\multicolumn{1}{|c|}{\textbf{F-score}} 
&\multicolumn{1}{|c|}{\textbf{Prec}}	&\multicolumn{1}{|c|}{\textbf{Rec}} &\multicolumn{1}{|c|}{\textbf{F-score}}
&\multicolumn{1}{|c|}{\textbf{Prec}}	&\multicolumn{1}{|c|}{\textbf{Rec}} &\multicolumn{1}{|c|}{\textbf{F-score}}\\ 
\hline
(1)& 0.6515 & 0.6220 & 0.6364 & 0.5464 & 0.6866 & 0.6085 & 0.6728 & 0.6214 & 0.6461\\
(2)& 0.5972 & 0.7454 & 0.6631 & 0.6119 & 0.7377 & 0.6689 & 0.6504 & 0.7261 & \textbf{0.6862}\\
(3)& 0.6193 & 0.6927 & 0.6540 & 0.6139 & 0.6575 & 0.6350 & 0.6738 & 0.6518 & 0.6626\\
\hline
Average & 0.6227 & 0.6867 & 0.6512 & 0.5907 & 0.6939 & 0.6375 & 0.6657 & 0.6665 & 0.6650\\
\hline
\hline
\multicolumn{1}{|c|}{\textbf{DrugBank}}	&\multicolumn{1}{|c|}{\textbf{Prec}}	&\multicolumn{1}{|c|}{\textbf{Rec}} &\multicolumn{1}{|c|}{\textbf{F-score}} 
&\multicolumn{1}{|c|}{\textbf{Prec}}	&\multicolumn{1}{|c|}{\textbf{Rec}} &\multicolumn{1}{|c|}{\textbf{F-score}}
&\multicolumn{1}{|c|}{\textbf{Prec}}	&\multicolumn{1}{|c|}{\textbf{Rec}} &\multicolumn{1}{|c|}{\textbf{F-score}}\\ 
\hline
(1) & 0.6973 & 0.6107 & 0.6512 & 0.6952 & 0.5847 & 0.6352 & 0.6081 & 0.6036 & 0.6059\\
(2) & 0.6961 & 0.6736 & \textbf{0.6847} & 0.6937 & 0.6479 & 0.6700 & 0.6836 & 0.6768 & 0.6802\\
(3) & 0.6976 & 0.6193 & 0.6561 & 0.6968 & 0.5929 & 0.6406 & 0.6033 & 0.6050 & 0.6042\\
\hline
Average & 0.6970 & 0.6345 & 0.664 & 0.6952 & 0.6085 & 0.6486 & 0.6317 & 0.6285 & 0.6301\\
\hline
\hline
\end{tabular}
 \end{table}

\subsubsection{LSTM Performance}
\label{LSTM_Performance}

\begin{figure}
\centering
\includegraphics[scale=0.7]{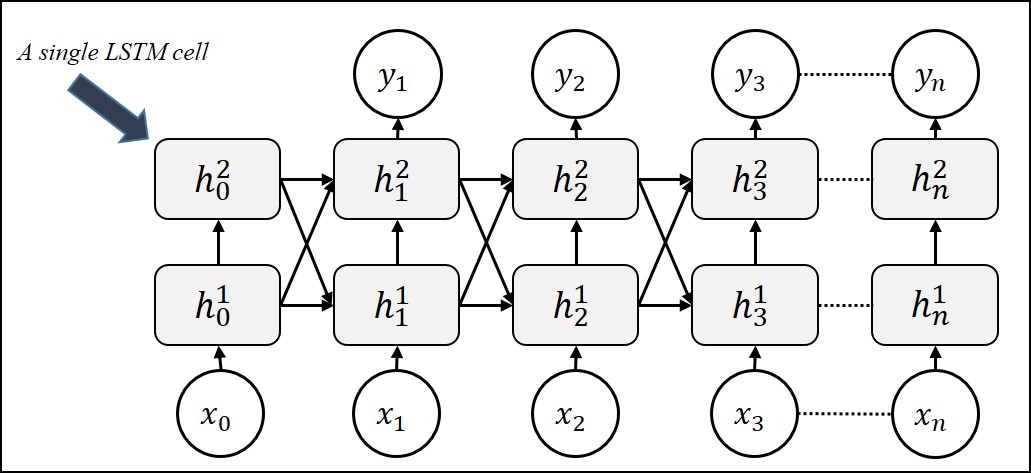}
\caption{The Global LSTM Network}
\label{fig:lstm_global}
\end{figure}

The global LSTM network used is presented in Figure \ref{fig:lstm_global}. Each single LSTM block consists of two stacked hidden layers, one input node with each input dimension is 200 as described in Subsection \ref{third_technique}. All hidden layers are fully connected. We used sigmoid as an output activation function, which is the most suitable for binary classification. We implemented a peepholes connection LSTM variant where its gate layers look at the cell state \cite{Fel}. In addition to implementing the peepholes connection, we also use a coupled of forget and input gates. The detail single LSTM architecture and its each gate formula computation can be referred in \cite{Olah2015}. 

The LSTM experiments were implemented with several different parameter settings. Their results presented in this section are the best among all our experiments. Each input data consists of two components, its word vector value and its Euclidian distance to the previous input data. In treating both input data components, we adapt the Adding Problem Experiment as presented in \cite{Hochreiter}. We use the Jannlab tools \cite{otte_jannlab_2013} with some modifications in the part of entry to conform with our data settings. 

The best achieved performance is obtained with LSTM block architecture of one node input layer, two nodes hidden layer, and one node output layer. The used parameters are: learning rate = 0.001, momentum = 0.9, and epoch = 30, input dimension = 200, and with the time sequence frame set to 2. The complete treatment of drug sentence as a sequence both in representation and recognition, to extract the drug name entities, is the best technique, as shown by F-score performance in Table \ref{table:fscore_LSTM}.

\begin{table}[h!]
\centering
\caption{The F-score performance of third data representation technique with RNN-LSTM}
\label{table:fscore_LSTM}
\begin{tabular}{| l | r | r | r |} 
\hline
\multicolumn{1}{|c|}{}& 
\multicolumn{1}{|c|}{\textbf{Prec}}& 
\multicolumn{1}{|c|}{\textbf{Rec}}&
\multicolumn{1}{|c|}{\textbf{Fscore}} \\
\hline
MedLine&1&0.6474&0.7859\\
DrugBank&1&0.8921&0.9430\\
\hline
Average&{}&{}&0.8645\\
\hline
\end{tabular}
\end{table}

As described in previous work section, there are many approaches related to drug extraction have been proposed. Most of them utilize certain external knowledge to achieve the extraction objective. The Table \ref{table:fscore_compared} summarizes their F-score performance. Among the state of the arts, our third data representation technique applied to the LSTM model is outperforming. Also, our proposed method does not require any external knowledge.

\begin{table}[h!]
\small
\caption{The F-score performance compared to the state of the art}
\label{table:fscore_compared}
\begin{tabular}{| l | r |l|} 
\hline
\multicolumn{1}{|c|}{\textbf{Approach}}& 
\multicolumn{1}{|c|}{\textbf{F-score}}
&\multicolumn{1}{|c|}{\textbf{Remark}} \\
Approach        &F-score    &Remark\\
\hline
\hline
The Best of SemEval 2013\cite{Segura-Bedmar2013}    &0.7150&-\\
\cite{Grego2013}            &0.5700   &With external knowledge, ChEBI\\
\cite{Segura-bedmar2015}+Wiki&0.7200   &With external knowledge, DINTO\\
\cite{Ben2015}         &0.7200   &Additional feature, BIO\\
\cite{Bjorne2013}           &0.6000    &Single token only\\
MLP-SentenceSequence+Wiki(average)/Ours&0.6580  &Without external knowledge\\
DBN-SentenceSequence+Wiki(average)/Ours&0.6430  &Without external knowledge\\
SAE-SentenceSequence+Wiki(average)/Ours&0.6480  &Without external knowledge\\
\multirow{2}{*}{LSTM-AllSentenceSequence+Wiki+}&
\multirow{2}{*}{0.8645}&
\multirow{2}{*}{Without external knowledge}\\
\multirow{2}{*}{EuclidianDistance(average)/Ours}&
\multirow{2}{*}{}&
\multirow{2}{*}{}\\
\multirow{2}{*}{}&
\multirow{2}{*}{}&
\multirow{2}{*}{}\\
\hline
\end{tabular}
\end{table}

\subsection{Drug Label Dataset Performance}
\label{Drug_Label_Dataset_Performance}

As additional experiment, we also use Indonesian language drug label corpus to evaluate the method's performance. Regarding the Indonesian drug label, we could not found any certain external knowledge that can be used to assist the extraction of the drug name contained in the drug label. In the presence of this hinderance, we found our proposed method is more suitable than any other previous approaches. As the drug label texts are collected from various sites of drug distributors, producers, and government regulators; it does not clearly contain training data and testing data as in DrugBanks or Medline datasets. The other characteristics of these texts are the more structured sentences contained in the data. Although the texts are coming from various sources, all of them are similar kind of document (the drug label that might be generated by machine). After the data cleaning step (HTML tag removal, etc.), we annotated the dataset manually. The total quantity of dataset after performing the data representation step, as described in Subsection \ref{data_representation}, is 1.046.200. In this experiment, we perform 10 times cross-validation scenario by  randomly selecting 80\% data for the training data and 20\% data for testing.

The experimental result for drug label dataset shows that all of the candidate selection scenarios provide excellent F-score (above 0.9). The excellent F-score performance is probably due to the more structured sentences in those texts. The best result of those ten experiments are presented in Table \ref{table:fscore_druglabel}.

\subsection{Choosing The Best Scenario}
\label{Choosing_the_best_scenario}
In the first and second experiments, we studied various experiment scenarios, which involves three investigated parameters: additional Wiki source, data representation techniques, and drug target candidate selection. In general, the Wiki addition contributes in improving the F-score performance. The additional source in word2vec training enhances the quality of the resulted word2vec. Through the addition of common words, from Wiki, the difference between the common words and the uncommon words i.e drug name becomes greater (better distinguishing power). 

One problem in mining drug name entity from medical text is the imbalanced quantity between drug token and other tokens \cite{Segura-Bedmar2013}. Also, the targeted drug entities are only a small part of the total tokens. Thus, majority of tokens are noise. In dealing with this problem, the second and third candidate selection scenarios show their contribution to reduce the quantity of noise. Since the possibility to extract the noises is reduced then the recall value and F-score value increase as well, as shown in the first and second experiments results. 

The third experiment which uses LSTM model does not applied the candidate selection scenario because the input dataset is treated as sentence sequence. So the input dataset can not be randomly divide (selected) as the tuple treatment in the first and second experiments. 

\begin{table}[h!]
\centering
\caption{The best performance of 10 executions on Drug Label Corpus}
\label{table:fscore_druglabel}
\begin{tabular}{| c | r | r | r |} 
\hline
\multicolumn{1}{|c|}{\textbf{Iteration}}	& \multicolumn{1}{|c|}{\textbf{Prec.}}	&\multicolumn{1}{|c|}{\textbf{Recall}}	&\multicolumn{1}{|c|}{\textbf{F-score}} \\
\hline
\hline
1	&0.9170	&0.9667	&0.9412\\
\hline
2	&0.8849	&0.9157	&0.9000\\
\hline
3	&0.9134	&0.9619	&0.9370\\
\hline
4	&0.9298	&0.9500	&0.9398\\
\hline
5	&0.9640	&0.9570	&0.9605\\
\hline
6	&0.8857	&0.9514	&0.9178\\
\hline
7	&0.9489	&0.9689	&0.9588\\
\hline
8	&0.9622	&0.9654	&0.9638\\
\hline
9	&0.9507	&0.9601	&0.9554\\
\hline
10	&0.9516	&0.9625	&0.9570\\
\hline
\hline
Average	&0.93081	&0.9560	&0.9431\\
\hline
Min	&0.8849	&0.9157	&0.9000\\
\hline
Max	&0.9640	&0.9689	&0.9638\\
\hline
\hline
\end{tabular}
\end{table}

\section{Conclusion $\&$ Future Works}
\label{Conclusion_Future_Works}
This study proposes a new approach in the data representation and classification to extract drug name entities contained in the sentences of medical text documents. The suggested approach solves the problem of multiple tokens for a single entity that remained unsolved in previous studies. This study also introduces some techniques to tackle the absence of specific external knowledge. Naturally, the words contained in the sentence follow a certain sequence patterns, i.e., the current word is conditioned by other previous words. Based on the sequence notion, the treatment of medical text sentences which apply the sequence NN model, gives better results. In this study, we presented three data representation techniques. The first and second techniques treat the sentence as a non-sequence pattern which is evaluated with the non-sequential NN classifier (MLP, DBN, SAE), whereas the third technique treats the sentences as a sequence to provide data that is used as the input of the sequential NN classifier i.e. LSTM. The performance of the application of LSTM models for the sequence data representation, with the average F-score being 0.8645, rendered the best result compared to the state of the art. 

Some opportunities to improve the performance of the proposed technique are still widely opened. The first step improvement can be the incorporation of additional handcrafted features - such as the words position, the use of capital case at the beginning of the word, the type of character - as also used in the previous studies \cite{Segura-bedmar2015,Boyce2012}. As presented in the MLP experiments for drug label document, the proposed methods achieved excellent performance when applied to the more structured text. Thus, the effort to make the sentence of the dataset, i.e., DrugBank and MedLine, to be more structured can also be elaborated. Regarding the LSTM model and the sequence data representation for the sentences of medical text, our future study will tackle the multiple entity extractions such as drug group, drug brand, and drug compounds. Another task that is potential to be solved with the LSTM model is the drug – drug interaction extraction. Our experiments also utilizes the Euclidean distance measure in addition to the word2vec features. Such addition gives a good F-score performance. The significance of embedding the Euclidean distance features, however, needs to be explored further.  

\section{Acknowledgment}
\label{Acknowledgment}
This work is supported by Higher Education Science and Technology Development Grant funded Indonesia Ministry of Research and Higher Education Contract No. 1004/UN2.R12/HKP.05.00/2016
\label{subsec1}

\section{Conflict of Interests}
\label{Conflict_of_Interests}
The authors declare that there is no conflict of interest regarding the publication of this paper.
\label{subsec1}


\bibliography{library}

\bibliographystyle{abbrv}

\end{document}